\documentclass{article}
\usepackage{spconf,amsmath,graphicx,amsfonts,hyperref}

\title{MODELING INTER-SPEAKER RELATIONSHIP IN XLNET FOR CONTEXTUAL SPOKEN LANGUAGE UNDERSTANDING}
\name{Jonggu Kim and Jong-Hyeok Lee}
\address{Department of Computer Science and Engineering,\\
	Pohang University of Science and Technology (POSTECH),\\
	77 Cheongam-Ro, Pohang, Gyeongbuk, Republic of Korea 37673}
\begin{document}
%
\maketitle
\begin{abstract}
We propose two methods to capture relevant history information in a multi-turn dialogue by modeling inter-speaker relationship for spoken language understanding (SLU). Our methods are tailored for and therefore compatible with XLNet, which is a state-of-the-art pretrained model, so we verified our models built on the top of XLNet. In our experiments, all models achieved higher accuracy than state-of-the-art contextual SLU models on two benchmark datasets. Analysis on the results demonstrated that the proposed methods are effective to improve SLU accuracy of XLNet. These methods to identify important dialogue history will be useful to alleviate ambiguity in SLU of the current utterance.
\end{abstract}
\section{Introduction}
\label{sec:intro}
Spoken language understanding (SLU) considers the context of a given utterance to improve SLU accuracy. For this purpose, speaker roles of historical utterances are also considered, to improve context-summarization methods: an encoder based on a recurrent neural network (RNN) for each speaker is used in encoding his or her own content separately \cite{Chi:IJCNLP17,Chen:ASRU17,Su:NAACL18,Su:ICASSP19,Kim:NAACL19}, and an utterance-level speaker-embedding vector that indicates a speaker of the utterance is used in utterance-level attention-based summarization methods \cite{Kim:Comp20}.

We propose two methods to model inter-speaker relationships for contextual SLU, which is tailored for and easily integrated with XLNet \cite{Yang:arxiv19}, a state-of-the-art pretrained model. Recently, pretrained models based on Transformer \cite{Vaswani:NIPS17} such as GPT \cite{Radford:18}, BERT \cite{Devlin:NAACL19} and XLNet \cite{Yang:arxiv19} achieve state-of-the-art accuracies in many fields of natural language processing (NLP) after being finetuned on their tasks. In this paper, we integrate the proposed methods with XLNet and verify them.

We explore effective ways for the model to consider historical utterances of a speaker important to spoken language understanding of the current utterance. Specifically, we investigate whether a ``speaker symbol token'' which indicates the speaker of the following utterance can improve accuracy of XLNet in multi-turn dialogues. Additionally, we propose ``relative speaker attention'', which provides scores to attention layers by evaluating whether speakers of two tokens are the same.

We conducted an experiment on two benchmark datasets, the fourth dialog state tracking challenge (DSTC 4) \cite{Kim:IWSDS16} and Loqui \cite{Passonneau:14}. In the experiment, all our models achieved state-of-the-art F1 scores on both datasets. To examine the effect of context length in our models, we conducted additional experiments and provided analysis of the results. The analysis of experiments demonstrated that the proposed methods are effective in improving SLU accuracy.

Our contributions are summarized as follows:
\begin{itemize}
\item We propose two methods of modeling inter-speaker relationship, speaker symbol token and relative speaker attention. We further integrate them with XLNet.
\item The proposed models achieved state-of-the-art F1 scores on two benchmark datasets, DSTC 4 and Loqui.
\end{itemize}

\begin{figure*}[t]
  \centering
  \noindent
  \includegraphics[width=\linewidth]{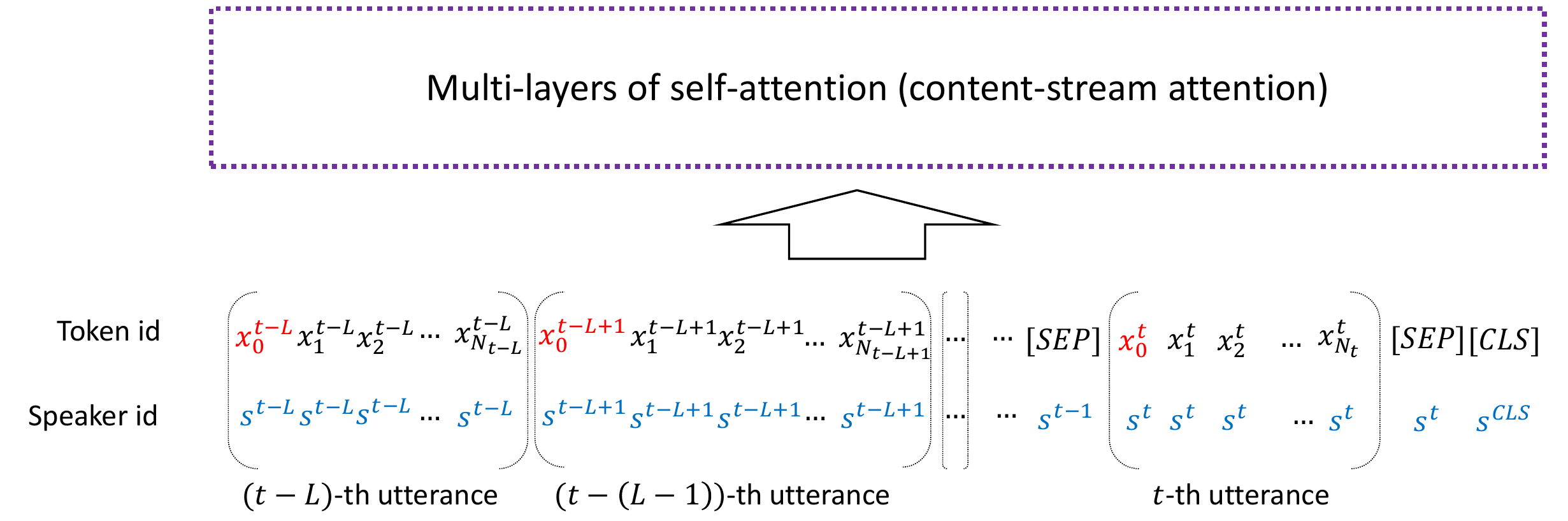}
  \caption{Inputs of the models for spoken language understanding of the current (\(t\)-th) utterance. \(L\) is the length of history our model considers, which means \((t-L)\) to \((t-1)\)-th utterances are in previous turns. \(N_{[.]}\) is the number of tokens of \([.]\)-th utterance. Token ids in red are for ``speaker symbol token'' and speaker ids in blue are for ``relative speaker attention''. A speaker symbol token indicates a speaker of the following utterance. We assign the speaker id of the last historical utterance to the first [SEP], the speaker id of the current utterance to the second [SEP] and a special speaker id to [CLS].}
  \label{fig:input_of_xlnet}
\end{figure*}

\section{RELATED WORK}
\label{sec:related_work}
Contextual models for SLU or for other fields of dialogue system have used recurrent neural networks (RNNs) to encode a sequence of utterances in time order \cite{Sordoni:CIKM15,Serban:AAAI17,Xu:ICASSP14,Chi:IJCNLP17,Bhargava:ICASSP13}. Usually, these models encode all inputs hierarchically at two levels: the token level and the utterance level. The models first encode all tokens to provide representations of each utterance, then use RNNs to encode the representations in time order at the utterance level.

Recent work focuses on capturing important histories to achieve understanding of the current utterance. Content-based attention models could not easily attend to important histories, so time-based attention models \cite{Chen:ASRU17,Su:NAACL18,Su:ICASSP19,Kim:NAACL19,Kim:Comp20} have been proposed. At the utterance level, time-based attention models consider distances of histories to focus attention on recent histories rather than earlier ones.

Previous approaches consider speaker roles, but only as auxiliary information. A different RNN-based encoder for each speaker is used to summarize the content \cite{Chi:IJCNLP17,Chen:ASRU17,Su:NAACL18,Su:ICASSP19,Kim:NAACL19}. A speaker-embedding vector that indicates the identity of the speaker of the utterance is directly added to its attention \cite{Kim:Comp20}. However, these previous methods are not easily integrated with other models like XLNet.

In this paper, we propose two speaker-modeling methods, ``speaker symbol token'' and ``relative speaker attention'', which are tailored for and integrated with XLNet. The resulting models achieved state-of-the-art F1 scores on DSTC 4 and Loqui datasets. Also, we verified effectiveness of the proposed methods.

\section{Proposed Method}
\label{sec:proposed_method}

\subsection{Framework}
\label{subsec:framework}
XLNet is an autoregressive model that has been trained to maximize the expected log likelihood of a text sequence with respect to all possible permutations of the factorization order. The objective used for pretraining is written as:

\begin{flalign}
  \max_{\theta} \mathbb{E}_{z\sim \mathcal{Z}_T}\Bigg{[} \sum_{t =1}^T \log p_\theta(x_{z_t}\mid x_{z_{<t}}) \Bigg{]},
  \label{eq:objective}
\end{flalign}
where \(\mathcal{Z}_T\) is the set of possible permutations of the \(T\) index sequence \([1, 2, ...,T]\), \(z_t\) is the \(t\)-th element and \(z_{<t}\) is the first \(t-1\) elements of a permutation \(z \in \mathcal{Z}_T\) \cite{Yang:arxiv19}.

This objective is expected to induce XLNet to capture bidirectional context for each token. Permutations are performed using attention masks, and they make direct or indirect links among all tokens under permutation orders.

For pretraining, the architecture consists of stacked layers of query-stream attention and content-stream attention. The query-stream attention starts with \(g_i^0 = w\) (a trainable vector) and the content-stream attention starts with \(h_i^0 = e(x_i)\) (an embedding vector of \(x_i\)). Then, for each layer \(m = 1, ... , M\), content-stream hidden representation \(h_{z_t}^m\) and query-stream hidden representation \(g_{z_t}^m\) are computed as:

\begin{flalign}
  g_{z_t}^m = \operatorname{Attention}(Q=g_{z_t}^{(m-1)}, K=h_{z_{<t}}^{(m-1)}, V=h_{z_{<t}}^{(m-1)}), \\
  h_{z_t}^m = \operatorname{Attention}(Q=h_{z_t}^{(m-1)}, K=h_{z_{\leq{t}}}^{(m-1)}, V=h_{z_{\leq{t}}}^{(m-1)}),
  \label{eq:attention}
\end{flalign}
where \(Q\), \(K\) and \(V\) stand respectively for the query, key and value in the attention mechanism \cite{Vaswani:NIPS17}. In reality, \(\operatorname{MultiHead(.)}\) is used, not \(\operatorname{Attention(.)}\), but we regard \(\operatorname{MultiHead(.)}\) as \(\operatorname{Attention(.)}\) for simplicity of explanation.

By finetuning on a specific task, the model is optimized for the task. The finetuning process considers only content-stream attention layers in which permutation masks are not used. Specifically, for each layer \(m = 1, ..., M\), the \(i\)-th content-stream hidden representation \(h_{i}^m\) is computed by two consecutive layers: an attention layer and position-wise feed-forward layer. In an attention layer, content-based attention score \(a_{ij}^{cont}\), position-based attention score \(a_{ij}^{pos}\) and segment-based attention score \(a_{ij}^{seg}\) are computed and merged to obtain an aggregate attention score \(a_{ij}^{total}\):

\begin{flalign}
  a_{ij}^{cont} = (q_i+b^{cont})^Tk_j^{cont}, \\
  a_{ij}^{pos} = (q_i+b^{pos})^Tk_j^{pos}, \\
  a_{ij}^{seg} = (q_i+b^{seg})^Tk_j^{seg}, \\
  a_{ij}^{total} = a_{ij}^{cont} + a_{ij}^{pos} + a_{ij}^{seg},
  \label{eq:attention_score}
\end{flalign}
where \(q_i = h_i^{m-1}W^Q\), \(k_j^{[.]} = h_j^{m-1}W^{[.]}\), and \(W^Q\) and \(W^{[.]}\) are weight matrices and \(b^{[.]}\)s are bias vectors.

\(a_{ij}^{total}\) is used in weighted sum of \(v_j\) to obtain \(\hat{h}_i^m\) after a softmax operation is applied as:

\begin{flalign}
  p_i = \operatorname{Softmax}(a_i^{total}), \\
  \hat{h}_i^m = \sum_j p_{ij}v_j,
  \label{eq:attention_repr}
\end{flalign}
where \(v_j = h_j^{m-1}W^{V}\) and \(W^{V}\) is a weight matrix.

Then \(\hat{h}_i^m\) is fed to a position-wise feed-forward network and added to its result

\begin{flalign}
  h_i^m = \hat{h}_i^m + \operatorname{PosFF}(\hat{h}_i^m).
  \label{eq:posff}
\end{flalign}

For a text-level classification task like SLU, the \(M\)-th layer, \(T\)-th query representation \(q_{T}^M\) is used to obtain

\begin{flalign}
  \hat{y} = \operatorname{Softmax}(\operatorname{FF}(q_{T}^M)),
  \label{eq:prob}
\end{flalign}
where \(\operatorname{FF}(.)\) is a feed-forward layer.

The objective of finetuning on the task is to optimize the conditional probability of labels, \(p(\hat{y}|x)\), by minimizing the cross-entropy loss.

\subsection{Modeling Inter-Speaker Relationship}
\label{Modeling Inter-Speaker Relationship}
We first propose to add a speaker symbol token at the front of each utterance (Fig. \ref{fig:input_of_xlnet}, tokens \(x_0\) in red). For DSTC 4, we add ``t:'' to indicate that the following utterance is by a tourist and ``g:'' to indicate that the following utterance is by a guide; in practice this token will be tokenized into two subtokens by a SentencePiece tokenizer \cite{Kudo:EMNLP18}. Similarly, we use two symbolic tokens at the front of each utterance for Loqui. This idea is inspired by BERT-based question answering models \cite{Reddy:TACL19}.

We additionally propose relative speaker attention that uses relative speaker embeddings to generate additional attention scores for self-attention. Relative speaker embeddings represent whether speakers of two different tokens are the same. Given two tokens \(x_i\) and \(x_j\), if they are of the same speaker, we obtain a relative speaker index \(s_{ij} = 1\) from their corresponding speaker ids \(s_i\) and \(s_j\) (Fig. \ref{fig:input_of_xlnet}, speaker ids \(s\) in blue). Otherwise, we obtain a relative speaker index \(s_{ij} = 0\). This index value is changed to a relative speaker key-vector \(k_j^{spk}\) by an embedding matrix, then used to compute a relative speaker attention score

\begin{flalign}
  a_{ij}^{spk} = (q_i+b^{spk})^Tk_j^{spk}.
\end{flalign}

We add \(a_{ij}^{spk}\) to the aggregate attention score \(a_{ij}^{total}\) to compute a new one. In other words, Eq. \ref{eq:attention_score} can be rewritten as:

\begin{flalign}
  a_{ij}^{total} = a_{ij}^{cont} + a_{ij}^{pos} + a_{ij}^{seg} + a_{ij}^{spk}.
  \label{eq:new_attention_score}
\end{flalign}

\section{Experiment}
\label{sec:experiment}

\subsection{Setting}
\label{ssec:setting}
We finetuned and used pretrained XLNet-Base (\url{https://github.com/zihangdai/xlnet}) as a baseline model. XLNet-Base has 12 layers, 768 hidden dimensions and 12 heads as a baseline model. In the experiment, XLNet-Base used the current-speaker token \(x_0^t\) because it is not a part of inter-speaker modeling, but a cause of improvement by itself. We used learning rate of \(1.0^{-5}\), batch size of 8, with 10,000 training steps, 1,000 warmup steps, and Adam optimizer \cite{Kingma:ICLR15} when finetuning all models. As an evaluation metric, we used the F1 score that is the harmonic mean of precision and recall. We used 14 recent histories (recent seven from tourist and recent seven from guide) with segment ``A''. For fair comparison with state-of-the-art SLU models, we used a ground-truth intent label of each historical utterance, not tokens of the utterance. We used tokens of a current utterance with segment ``B''.

We used the of DSTC 4 \cite{Kim:IWSDS16} and Loqui \cite{Passonneau:14} datasets in experiments. DSTC 4 consists of 31,034 utterances of dialogues on tourist information of Singapore. Two speakers (a tourist and a guide) appear in the dialogues. The utterances are divided into 35 sessions. Each utterance has a speech act and its attributes. We divided the 35 sessions into the same training dataset, the same test dataset and the same validation dataset using the same indices as in the DSTC 4 competition.

Loqui consists of about 8,200 utterances of telephone transactions of New York City’s Andrew Heiskell Braille \& Talking Book Library. Two speakers (a librarian and a patron) appear in the transactions. The utterances are divided into 82 transactions. The utterances are annotated with dialogue acts. Among the 82 transactions, we used the same 62 training dataset, the same 10 test dataset and the same 6 validation dataset as in \cite{Kim:Comp20}.

\begin{figure}[t]
  \centering
  \noindent
  \includegraphics[width=\linewidth]{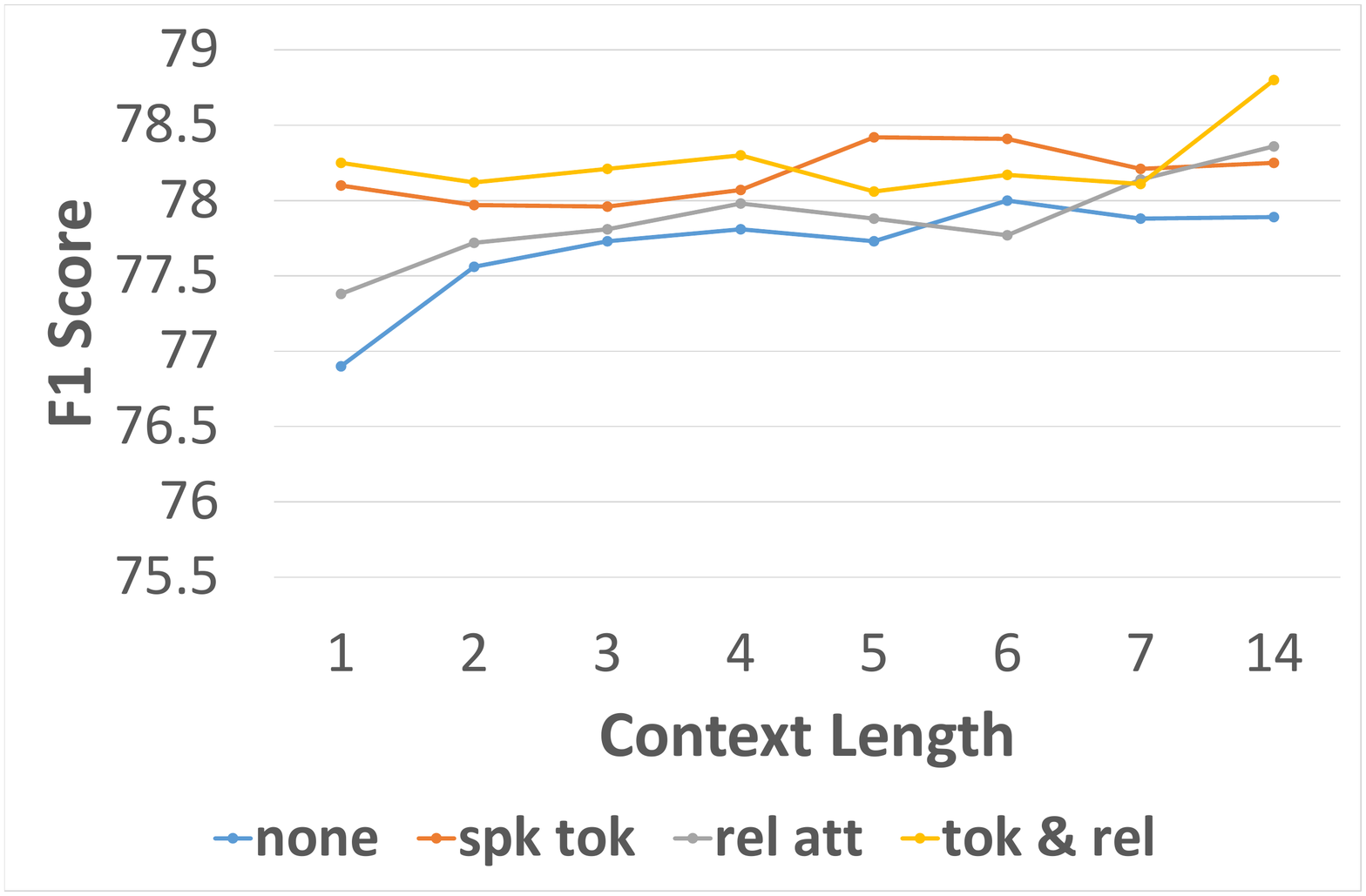}
  \caption{F1 scores of the models with different context lengths on the validation dataset of DSTC 4. none: XLNet-Base; spk tok: XLNet-Base with Speaker Symbol Token; rel att: XLNet-Base with Relative Speaker Attention; tok \& rel: XLNet-Base with Speaker Symbol Token \& Relative Speaker Attention.}
  \label{fig:chart_f1}
\end{figure}

We trained and tested each model five times and report the median of the F1 scores (Table \ref{table:entire_result}). For other models, results that are not in the original papers are taken from \cite{Kim:Comp20}. As an alternative evaluation, we obtained F1 scores (Fig. \ref{fig:chart_f1}) by running each model once for the number of history turns.

\begin{table*}
  \begin{center}
    \caption{F1 score of state-of-the-art models and the proposed models on the test dataset and the validation dataset of DSTC 4 and Loqui.}
    \begin{tabular}{clcccc}
    \textbf{\#} &\textbf{Model} & \multicolumn{2}{c}{\textbf{DSTC 4}} & \multicolumn{2}{c}{\textbf{Loqui}} \\
    & & Test & Valid & Test & Valid \\
    \hline
    1&Convex Time \cite{Chen:ASRU17} & 74.6  & - & - & - \\
    2&Universal Cont + Time \cite{Su:NAACL18} & 74.40 & 73.87 & 67.83 & 54.88 \\
    3&Context-Sensitive Time \cite{Su:ICASSP19} & 74.20 & 73.44 & 67.22 & 55.63 \\
    4&Decay-Function-Free Cont + Time  \cite{Kim:NAACL19}& 76.11 & 75.74 & 69.12 & 57.46 \\
    5&Speaker-Informed Hybrid Cont + Time \cite{Kim:Comp20} & 76.86 & 76.71 & 72.74 & 62.74 \\
    \hline
    6&XLNet-Base & 77.84 & 78.34 & 73.77 & 62.61 \\
    7&  + Speaker Symbol Token & 78.12 & 78.61 & 72.77 & \textbf{63.45} \\
    8&  + Relative Speaker Attention & 77.60 & 78.36 & \textbf{74.90} & 62.56 \\
    9&  + Speaker Symbol Token \& Relative Speaker Attention & \textbf{78.24} & \textbf{78.82} & 72.40 & 62.85
    \label{table:entire_result}
    \end{tabular}
  \end{center}
\end{table*}

\subsection{Result and Analysis}
\label{ssec:result_and_analysis}
All XLNet-based models achieved significantly higher F1 scores than state-of-the-art models (Table \ref{table:entire_result}). XLNet-Base with both speaker-modeling methods (Speaker Symbol Token and Relative Speaker Attention) achieved the best F1 scores on the test and validation datasets of DSTC 4 (row 9). On DSTC 4 datasets, XLNet-Base with Speaker Symbol Token (row 7) was always better than XLNet-Base (row 6). When Relative Speaker Attention was added to row 7 (row 9), the model achieved the highest F1 scores on both test and validation datasets. However, adding Relative Speaker Attention did not always improve F1 scores (row 8).

\vfill\pagebreak

XLNet-Base with Relative Speaker Attention achieved the best F1 score on the test dataset of Loqui (row 8) and with Speaker Symbol Token achieved the best F1 score on the validation dataset of Loqui (row 7).

We also measured F1 scores of the models for different amounts of conversation history (Fig. \ref{fig:chart_f1}). For all models, F1 score increased with increase in the number of histories used. This tendency may be caused by positive effect of token-level position embeddings of XLNet: the model may not need any time-decay functions to attend to recent historical utterances; \textit{cf.} \cite{Chen:ASRU17,Su:NAACL18,Su:ICASSP19,Kim:NAACL19,Kim:Comp20}. Also, we found XLNet-Base achieved the lowest F1 scores in most different context lengths. This result demonstrates that the proposed methods improved accuracy of XLNet regardless of the number of given histories. XLNet-Base with Speaker Symbol Token (and Relative Speaker Attention) for most context lengths achieved good F1 scores. We found that Relative Speaker Attention tends to boost F1 scores of the model, whereas XLNet-Base cannot break its upper bound when 14 turns are given.

\section{Conclusion}
\label{sec:conclusion}
We proposed two inter-speaker modeling methods, speaker symbol token and relative speaker attention. We integrated the proposed methods with XLNet, a state-of-the-art pretrained model, and therefore we proposed combined models. To verify effectiveness of the models, we conducted experiments on two benchmark datasets, DSTC 4 and Loqui. In the experiments, all models achieved state-of-the-art F1 scores on both datasets. We provided analysis on results of the experiments. The result of analysis demonstrated that the proposed methods of modeling inter-speaker relationship can encourage improvement of the accuracy of XLNet.

\vfill\pagebreak

\bibliographystyle{IEEEbib}
\bibliography{Template_mine}

\end{document}